# The MediaSpin Dataset: Post-Publication News Headline Edits Annotated for Media Bias

**Preetika Verma**[1,2] **, Kokil Jaidka**[2,3]
[1]Carnegie Mellon University, USA
[2] Department of Communications and New Media, National University of Singapore
[3]NUS Centre for Trusted Internet and Community, National University of Singapore, Singapore
jaidka@nus.edu.sg

### Abstract

We present MediaSpin, a large-scale language resource capturing how major news outlets modify headlines after publication, and MediaSpin-in-the-Wild, a complementary dataset linking these revised headlines to their downstream engagement on social media. The increasing editability of online news headlines offers new opportunities to study linguistic framing and bias through the lens of editorial revisions. The dataset contains 78,910 headline pairs annotated for 13 types of media bias, grounded in established media-bias taxonomies, covering both subjective (e.g., sensationalism, spin) and objective (e.g., omission, slant) forms, with annotation conducted through a human-supervised large-language-model pipeline with expert validation and quality control. We describe the annotation schema and demonstrate three downstream applications: (1) cross-national analysis of how country references are added or removed during editing, (2) transformer-based bias classification at both binary and fine-grained levels, and (3) behavioral analysis of biased headlines on X (Twitter) using 180,786 news-related tweets from 819 consenting users. The results reveal regional asymmetries in representational framing, measurable linguistic markers, and consistently higher engagement with biased content. MediaSpin and MediaSpin-in-the-Wild together provide a reproducible benchmark for bias detection and the study of editorial and behavioral dynamics in contemporary media ecosystems.

## Introduction

Published news articles play a significant role in shaping public opinion. The increasing editability of online news content, facilitated by social media platforms, introduces new affordances for dynamic and adaptive news framing. Edits to news headlines can refocus audience attention, add or remove emotional language, and shift the framing of events in subtle yet impactful ways. This phenomenon of dynamic news framing, where headlines are revised post-publication, remains underexplored in prior research.

Although the literature on news framing dates back several decades (Tewksbury and Scheufele 2019), news editors may intentionally or accidentally introduce or remove biases when revising published news reports. These edits can emphasize frames of salience, acknowledge key actors or context, or update unfolding events (Tewksbury and Scheufele 2019). News edits can also involve the inclusion or removal of affective content or critical details relevant to the news report's agenda-setting and framing roles.

Prior research on editing practices, such as in Wikipedia (Jaidka et al. 2021), provides useful frameworks to characterize editing intentions, like clarification, elaboration, specification, and enforcing neutrality. However, these frameworks are insufficient for categorizing the linguistic bias embedded in media content. Simultaneously, analyses of media bias by platforms like AllSides[1] often overlook the editorial decisions reflected in subtle linguistic shifts.

| Type of Bias | Edited Headline | Original Headline | Change |
|---|---|---|---|
| Unsubstantiated Claims | 36 Bodies Are Found at Manila Casino | Terrorism Fear at Manila Casino May Have Been Caused by Irate Gambler Hours After Fires | Removal |
| Opinion Statements Presented as Fact | Central Bank Sets Bond Plan Meant to Ease Euro Debt Peril | Central Bank to Snap Up Debt, Saying, "Euro Is Irreversible" | Addition |
| Sensationalism/ Emotionalism | Buying at a Used Car Auction? What You Don't Know Could Kill You | At Auction, Vehicles With Fatal Flaws | Addition |
| Mind Reading | Electronic Brain Attacks Drove Shooter to Violence, F.B.I. Says | Navy Yard Shooter Believed Mind Was Under Attack, Official Says | Removal |
| Flawed Logic | Ex-Friend of Robert Durst Testifies Against Him in Advance of Murder Trial | Lawyer Says Robert Durst's Former Friend Concocted His Story of a Confession | Addition |
| Omission of Source Attribution | Close to a Million Could Be Tested for the Coronavirus This Week, Health Official Says | Close to a Million in U.S. Could Be Tested for Coronavirus This Week | Removal |
| Word Choice | Ukraine, Eye on Convoy, Says It Hit Military Vehicles Coming From Russia | Russian Armed Vehicles Destroyed After Crossing Border, Ukraine Says | Addition |

Table 1. Examples from the MediaSpin dataset

Studying linguistic biases in news editing practices provides insights into the prevailing social and political attitudes of media outlets and the broader sociopolitical context. For example, Woo and Kim (2023) demonstrate how female par-liamentarians in South Korea are portrayed with more negative tones than their male counterparts. Such findings are increasingly relevant with the rise of large language models (LLMs), which can be leveraged to build real-time content moderation tools that detect biased news framing.

[1]https://www.allsides.com/

Moreover, Jiang et al. (2024) show how consumption of politically biased news about COVID-19 vaccines can shift people's stances on vaccination, illustrating the real-world impact of media framing.

To address these research gaps, our study presents the MediaSpin dataset[2] and a multidisciplinary approach that combines computational methods with social science theory. Drawing on established multi-dimensional taxonomies of media bias (Vargas et al. 2023; Spinde et al. 2023; Hamborg, Donnay, and Gipp 2020), we developed a labeled dataset of 78,910 headline pairs annotated with 13 types of media bias, spanning both subjective dimensions (e.g., sensationalism, spin, mudslinging, opinion presented as fact) and objective dimensions (e.g., omission, slant, unsubstantiated claims, flawed logic). These categories capture bias patterns introduced or removed through editorial decisions. We demonstrate the dataset's utility through three downstream applications: a cross-national analysis of how country references are systematically added or removed during headline editing, revealing geographic asymmetries in representational framing; transformer-based bias classification evaluating both binary and fine-grained prediction; and an examination of the downstream effects of media bias on selective news sharing through an analysis of over 180k tweets mentioning news handles.

## Related Work

### Framing and Media Bias in News Texts

News framing research shows that linguistic choices structure how audiences interpret political actors, issues, and events. Small lexical substitutions can shift evaluative meaning or invoke culturally salient frames (Hamborg 2023; Entman et al. 2004). Labeling choices such as "rioters" vs. "protesters" or "illegal" vs. "undocumented" immigrants systematically reflect ideological positioning (Card et al. 2015). Computational systems such as Newsalyze model these differences by identifying sentiment and frame-bearing lexical variation across outlets (Hamborg 2023).

Recent NLP work expands this view by revealing how framing emerges through connotation, implicit causality, and the allocation of agency (Sap et al. 2020). Media scholar-ship similarly documents metonymic shortcuts ("Asia," "the markets"), racialized framings, and geopolitical hierarchies (Amores, Arcila-Calderón, and González-de Garay 2020; Woo and Kim 2023). Multi-dimensional taxonomies of media bias highlight omission, attribution, emotionality, and fact–opinion blending as consequential dimensions for political attitudes and trust (Hamborg, Donnay, and Gipp 2020; Spinde et al. 2023).

[2]The Python and R package to annotate biases is available at https://github.com/kokiljaidka/mediaspin. The dataset is available at https://doi.org/10.7910/DVN/MOCQTZ.

**Research gap.** Nearly all computational approaches analyze final published headlines or articles. Although framing is theorized as the outcome of iterative editorial decisions, current corpora do not capture the revision process, making it difficult to study how subjective and objective biases arise through specific edits. Building on these multi-dimensional taxonomies, our annotation schema operationalizes 13 bias categories drawn from Spinde et al. (2023) and Hamborg, Donnay, and Gipp (2020), organized into subjective dimensions (sensationalism, spin, mudslinging, mind reading, subjective adjectives, word choice, opinion as fact) and objective dimensions (unsubstantiated claims, slant, flawed logic, omission, omission of source attribution, bias by story choice and placement). For headline-level annotation, *bias by story choice and placement* is operationalized as changes in topical emphasis or the foregrounding of particular story angles through editorial revision, e.g., when a headline shifts from covering an economic event to framing it as a political confrontation, rather than physical placement on a page.

### Dynamic Editing, Revision Histories, and Editorial Intent

Work on revision histories in collaborative platforms shows that edits contain meaningful signals about stance, conflict, and epistemic negotiation (Bender et al. 2011). Sim-ilar patterns appear on social media, where message edits reflect rhetorical softening or strategic reframing (Glenski, Weninger, and Volkova 2018). In journalism, newsroom studies document frequent headline updates, corrections, and stealth edits prompted by breaking news, legal pressures, or audience optimization (Spangher et al. 2022; d'Andréa 2009). Research on media bias cautions against interpreting final text as evidence of editorial intent. Stylistic norms, routinized workflows, and production constraints can produce systematic slant absent deliberate ideological aims (Gentzkow and Shapiro 2006; Groseclose and Milyo 2005). Media sociology further shows that editors routinely compress or expand context depending on space or time pressure, making revision operations themselves analytically valuable.

**Research gap.** Existing revision datasets do not annotate *what types of bias* edits introduce or remove. No framework links additive or subtractive edits to categories such as sensationalism, omission, attribution loss, or geographic reframing. Thus, the editorial process behind bias remains largely unobservable.

### News Consumption Behaviors

Audience studies show that exposure to news is shaped by platform structure, following networks, and ideological clustering (Bakshy, Messing, and Adamic 2015). Behavioral signals can infer outlet-level slant at scale (Ribeiro et al. 2018), and variation in framing or issue emphasis across outlets produces diverging information environments (Lin, Bagrow, and Lazer 2011).

Affective and sensational language reliably increases visibility: headlines with emotional or dramatic phrasing elicit more reactions, shares, and comments (Berger and Milkman 2012).

**(a) Media Outlets**

| Outlet | Count | Bias | Cred. |
|---|---|---|---|
| Washington Post | 59,749 | Left-centre | High |
| Reuters | 11,822 | Least biased | High |
| Fox News | 5,042 | Right | Low |
| New York Times | 2,183 | Left-centre | High |
| Rebel | 114 | Right | Low |

**(b) Bias Label Distribution**

| Bias Type | Added | Removed | No Bias |
|---|---|---|---|
| Spin | 6,085 | 1,549 | 71,216 |
| Unsubstantiated Claims | 5,145 | 908 | 72,803 |
| Opinion as Fact | 6,000 | 1,302 | 71,559 |
| Sensationalism/Emotionalism | 6,673 | 768 | 71,436 |
| Mudslinging | 1,963 | 1,009 | 75,882 |
| Mind Reading | 279 | 71 | 78,537 |
| Slant | 33,669 | 3,242 | 41,895 |
| Flawed Logic | 20 | 2 | 62,491 |
| Bias by Omission | 4,413 | 9,442 | 64,981 |
| Omission of Source Attribution | 1,425 | 1,430 | 76,027 |
| Bias by Story Choice / Placement | 4,567 | 604 | 73,698 |
| Subjective Qualifying Adjectives | 19,089 | 1,633 | 58,102 |
| Word Choice | 17,893 | 3,550 | 54,731 |

Table 2. MediaSpin Dataset overview. (a) Outlet-level description with political bias and credibility scores (source: Robertson et al. (2018)); (b) Label distribution across bias types.

**Research gap.** Prior work evaluates engagement on static headlines or clickbait features, not on how edits alter engagement. We lack evidence on whether headlines that become more biased through revision behave differently from those biased at publication, or whether subjective and objective changes diffuse differently across platforms.

## Annotating News Edits

Our research builds upon the NewsEdits dataset (Spangher et al. 2022), which reports the text revision patterns across 1.2 million published articles from 22 media outlets, which have been revised nearly four times on average to yield a total of 4.6 million revisions. Expanding on their work, we developed the MediaSpin dataset that tracks word-level changes, and labels them to facilitate a better understanding of how linguistic changes contribute to the framing of biases.

### Data Collection and Annotation Process

- We selected headline edit pairs from the NewsEdits dataset for five English language media outlets span-ning the ideological and credibility spectrum according to Robertson et al. (2018): Fox News, New York Times, Washington Post, Reuters, and Rebel.
- The pairs were cleaned by removing punctuation and phrases (e.g., "| Fox News") and generating lists of inserted and removed words.
- We annotated editorial bias using an LLM pipeline, specifically GPT-3.5-turbo, following the prompt in the Appendix Figure 3.

Table 2b presents the label distribution.

Following this method, we observed headline updates in approximately 17% version pairs (376,944 after sampling 2 million pairs). Table 1 provides detailed examples from the dataset with explanations related to the annotated labels. Table 2a provides the number of edited news headlines per news outlet in our dataset.

### Human Validation

To validate the annotations, we sampled forty instances (twenty each for instances of biases added and removed) of each type of media bias, which were then independently reviewed and annotated by two co-authors. This resulted in a total of 509 instances, as in some cases there were fewer than forty samples present. Annotators agreed with the annotations on 432 out of 509 samples. The overall inter-annotator agreement (pairwise percentage agreement) was 84.9%, with a corresponding Cohen's $\kappa$ of 0.67, indicating substantial agreement beyond chance. The inter-annotator agreement on the subset where annotations indicated that edits *introduced* new biases was higher, at 87.7% ($\kappa = 0.72$), while agreement on annotations where bias was suspected to be removed was slightly lower at 82.0% ($\kappa = 0.61$).

**Per-category reliability.** Agreement varied across bias types. Subjective bias categories achieved near-perfect agreement: sensationalism (100%), spin (95%), mudslinging (97.5%), and subjective adjectives (100%). Objective categories showed more variability: omission (80%), unsubstantiated claims (75%), and flawed logic (72.5%), reflecting the greater inferential demands of these categories, which often require contextual knowledge beyond the headline pair itself.

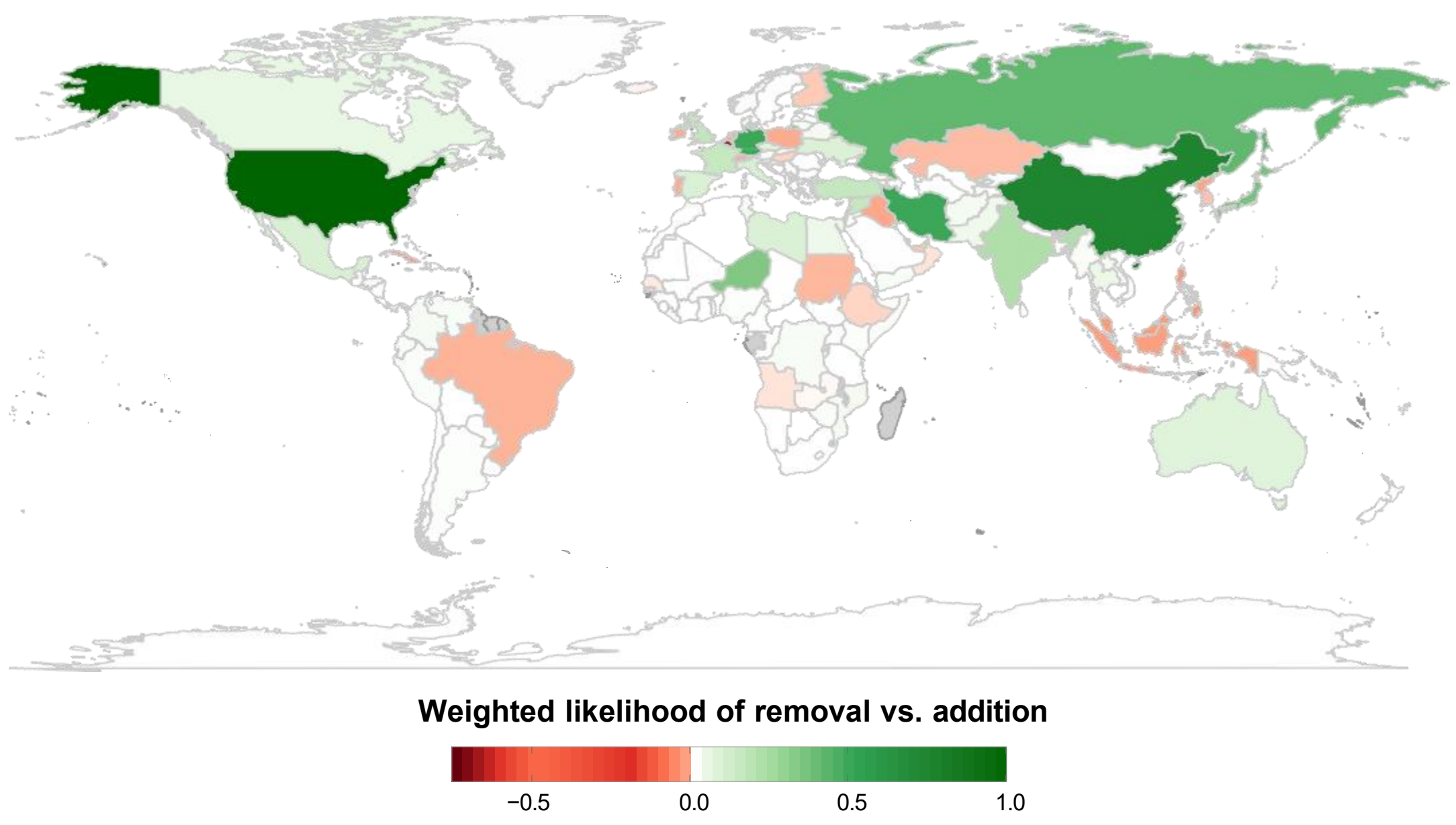

Figure 1: Weighted country-level likelihood of being added or removed. Green = more often added. Red = more often removed. White = mentioned in corpus but not edited. Gray = least mentioned in any headlines.

| Country | Added | Removed | Total | Score |
|---|---|---|---|---|
| *Most frequently added* | | | | |
| United States | 5025 | 4851 | 9876 | 1.000 |
| China | 720 | 556 | 1276 | 0.733 |
| Iran | 351 | 230 | 581 | 0.481 |
| Germany | 1699 | 1604 | 3303 | 0.481 |
| Russia | 501 | 406 | 907 | 0.404 |
| *Most frequently removed* | | | | |
| Belgium | 793 | 945 | 1738 | −0.709 |
| Philippines | 66 | 81 | 147 | −0.047 |
| Cuba | 45 | 59 | 104 | −0.041 |
| Malaysia | 50 | 63 | 113 | −0.038 |
| Netherlands | 85 | 95 | 180 | −0.032 |
| *Only removed, never added* | | | | |
| Iceland | 0 | 2 | 2 | −0.001 |
| Djibouti | 0 | 1 | 1 | <0.001 |
| Georgia | 0 | 1 | 1 | <0.001 |
| Palau | 0 | 1 | 1 | <0.001 |
| Palestinian Terr. | 0 | 1 | 1 | <0.001 |
| Trinidad & Tobago | 0 | 1 | 1 | <0.001 |

Table 3. Top five countries by frequency-weighted likelihood of being added or removed, and countries that are exclusively removed. Score denotes the normalized frequency-weighted score, ranging from −1 (predominantly removed) to +1 (predominantly added).

**Error analysis.** The GPT annotations showed inaccuracies primarily in objective bias categories: the model was sometimes more cautious than the human annotators, flag-ging headlines as insufficiently substantiated when annotators judged them acceptable. The most common error type was directional: GPT-3.5-turbo indicated an addition of bias when the edit actually removed bias. In fewer cases, bias was correctly identified but assigned to the wrong category.

**Limitations of single-LLM annotation.** Our pipeline relies on a single LLM (GPT-3.5-turbo). We acknowledge that using multiple LLMs and assessing inter-model agreement could reduce the influence of model-specific biases, and we recommend this as a direction for future validation work. The human validation reported here serves as a corrective check, but more extensive annotation with diverse models would strengthen reliability, particularly for the more ambiguous objective bias categories.

In the following sections, we demonstrate the usefulness of our dataset through three examples of its application. We also discuss more advanced applications and tools that can built on top of them.

## Downstream Tasks

### Cross-National Analysis of Editorial Changes

We extracted a place-country lookup from the headlines using GeoNamesCache and pycountry. Each place was then matched to its country using the OpenStreetMap Nominatim API, and the results were further cleaned by prompting GPT-4o-mini, followed by manual review.

To visualize geographic variation in the likelihood of country references being added or removed, we computed a frequency-weighted score for each country $i$:

$$\text{Score}_{\text{weighted},i} = \frac{(\text{Added}_i - \text{Removed}_i) \cdot \log(1+\text{Total}_i)}{\max_j \left| (\text{Added}_j - \text{Removed}_j) \cdot \log(1+\text{Total}_j) \right|}$$

This formulation rewards countries with higher total mention frequency while preserving the directional balance between additions and removals. The resulting normalized score ranges from $-1$ (predominantly removed) to $+1$ (predominantly added).

In Figure 1, countries never mentioned in any headline are rendered in grey, those mentioned but never edited appear in white, and scored countries are shaded from red (predominantly removed) to green (predominantly added). The visualization highlights clear regional asymmetries in representational dynamics. Table 3 lists the top five countries most frequently added and removed based on the weighted scores, alongside countries that are exclusively removed. The United States is most likely to be added, followed by China, Iran, Germany, and Russia: countries that dominate geopolitical news cycles and whose specificity is editorially foregrounded. Among removed countries, Belgium leads substantially, driven by EU-related headlines where "Belgium" is replaced with broader labels such as "EU" or "Europe," followed by the Philippines, Cuba, Malaysia, and the Netherlands: smaller nations whose specificity is more readily collapsed into regional or thematic framings.

A separate category of countries, including the Palestinian Territories, Djibouti, and Georgia, are exclusively removed and never added during editorial revision, representing cases where editorial processes only ever reduce their visibility. Although these cases are rare in absolute terms, they illustrate a form of representational asymmetry that frequency-weighted scoring necessarily underweights. A limitation of this approach is that the weighting favors countries with high mention counts; countries that appear infrequently but are systematically edited in one direction remain difficult to detect at scale. Additionally, 28 countries that appear in headlines are never subject to any editorial addition or removal (e.g., South Sudan, Eritrea, Bolivia, Western Sahara), suggesting a form of representational stasis in which these nations are present but never editorially foregrounded or backgrounded.

### Bias Classification

Next, we sought to evaluate whether the annotated dataset could support the development of a dynamic content suggestion tool. Understanding these biases is crucial for identifying patterns in media content and assessing potential editorial influences. To this end, we used the labeled dataset from the previous section to fine-tune and evaluate deep learning models for their predictive performance in bias classification. We categorized all the bias categories into two groups: subjective and objective, because these categories reflect distinct editorial choices that can influence audience perception differently. *Subjective biases* encompass linguistic features such as opinion statements presented as fact, sensationalism, mudslinging, mind reading, subjective qualifying adjectives, word choice, and spin. *Objective biases* include unsubstantiated claims, slant, flawed logic, bias by omission, omission of source attribution, and bias by story choice and placement.

For this analysis, we utilized 10,788 pairs of headlines labeled for the presence of objective bias, and 14,415 pairs labeled for subjective bias. To create a balanced dataset, we sampled an equivalent number of headline pairs from the remaining data. The headlines were labeled based on predefined linguistic markers indicative of each bias type.

**Classification setup** We chose DeBERTa-v3-small (He, Gao, and Chen 2023), a transformer-based model known for its efficiency and strong performance on text classifica-tion tasks. We employed a 70-30 train-test split, ensuring that the test set remained unseen during model training. The following were the finetuning hyperparameters:

- epochs = 10
- learning rate = 2e-5
- per_device_train_batch_size = 128
- per_device_eval_batch_size = 64
- weight decay = 0.01

Table 4 presents the predictive performance of DeBERTa-v3-small after fine-tuning on our dataset. The results demonstrate that while the model achieves relatively strong accuracy and minority-class F1 scores, particularly for subjective bias detection, the lower F1 scores for minority classes underscore the inherent challenge of detecting less frequently occurring bias patterns, particularly for objective bias. **Fine-grained 13-class classification.** To evaluate whether the full taxonomy can support multi-class prediction, we conducted a pilot experiment using the same DeBERTa-v3-small architecture on the 13 individual bias categories. Each category was treated as a separate binary classification task (one-vs-rest). Performance varied considerably across categories: subjective bias types with clearer lexical signals (e.g., sensationalism, F1 =

| Model | Subjective Bias | | Objective Bias | |
|---|---|---|---|---|
| | Acc. | Min. F1 | Acc. | Min. F1 |
| (He, Gao, and Chen 2023) | 0.773 | 0.758 | 0.774 | 0.761 |

Table 4. Predictive performance with DeBERTa-v3 for two binary classification tasks.

0.71; spin, F1 = 0.68) were more reliably detected, while objective categories requiring inferential reasoning (e.g., flawed logic, F1 = 0.48; omission, F1 = 0.52) proved substantially harder. These results confirm that the fine-grained taxonomy captures meaningful distinctions beyond the binary grouping, while also highlighting that objective bias detection remains an open challenge requiring richer contextual modeling.

**Linguistic insights** To identify the word choices associated with different forms of bias, we calculated the correlation between the term frequency-inverse document frequency (TF-IDF) scores of the editorialized words and their corresponding subjective and objective bias labels. Table 5 present rows shaded in green, indicating a statistically significant correlation with either subjective or objective bias, at the $p < 0.05$ level, after applying Benjamini-Hochberg correction for multiple comparisons. The analysis offers face validity to our dataset: for subjective bias, words such as *china* and *US* are common to both lists as they are associated with editorialized content in general. Words like *says* and *white* appeared more frequently in subjective contexts, suggesting a possible association with quotes and coverage of racism. The words *kills* and *dead* showed a slight negative correlation with objective bias, indicating that more objective headlines might use more direct and less emotionally charged terms for fatal events.

## Behavioral Analysis of Biased Headlines on X

To examine how editorial bias shapes downstream audience behavior, we introduce **MediaSpin-in-the-Wild**, a behavioral dataset linking the MediaSpin bias annotations to real-world news sharing on X (Twitter). MediaSpin-in-the-Wild contains 180,786 news-related tweets from 819 consenting American users, each annotated using the same 13-category subjective and objective bias taxonomy applied to headline edits. This paired design allows us to evaluate whether biased framings—whether affective (subjective) or evidentiary (objective)—are associated with systematic differences in engagement once headlines circulate on social platforms.

Figure 2 visualizes these engagement differences by comparing the average likes, replies, and retweets received by biased versus unbiased news content. The analysis leverages participants' public timelines collected as part of a broader survey on political discussion and news consumption.

**Participant recruitment** Data were collected through an online survey administered via Qualtrics. The sample consisted of 819 participants, skewed toward Democrats (65.9%), with 25.2% identifying as Republican and 8.9% as independent or other affiliations. The gender distribution was 58.7% male, 41.0% female, and 0.2% identifying otherwise.

**(a) Subjective Bias**

| Word | Count | Corr. |
|---|---|---|
| new | 2,935 | 0.10 |
| trump | 2,753 | 0.09 |
| us | 3,958 | 0.05 |
| trumps | 1,042 | 0.04 |
| attack | 1,189 | 0.04 |
| china | 968 | 0.04 |
| white | 696 | 0.04 |
| police | 1,434 | 0.04 |
| syria | 684 | 0.04 |
| live | 751 | −0.01 |
| updates | 716 | −0.01 |

**(b) Objective Bias**

| Word | Count | Corr. |
|---|---|---|
| trump | 2,753 | 0.12 |
| us | 3,958 | 0.10 |
| says | 3,826 | 0.09 |
| white | 696 | 0.06 |
| china | 968 | 0.05 |
| russia | 755 | 0.05 |
| police | 1,434 | 0.05 |
| state | 776 | 0.05 |
| kills | 770 | −0.01 |
| least | 778 | −0.01 |
| dead | 1,227 | −0.01 |

Table 5. Top words associated with (a) subjective and (b) objective bias (either added or removed). Shaded rows indicate statistical significance at $p < 0.05$ after Benjamini-Hochberg correction.

**Analytical approach** To examine whether media bias predicts engagement on X, we retrieved users' public timelines via the X API and extracted news-related retweets with their engagement metrics (likes, replies, and retweets). We restricted the analysis to the 180,786 tweets referencing known news headlines from the MediaSpin dataset. We then merged the two datasets to obtain bias labels for each tweet, classifying a tweet as subjectively or objectively biased if it exhibited at least one constituent category of the respective bias dimension; this yielded 97,719 subjectively biased and 41,883 objectively biased tweets.

Figure 2 presents log-scaled mean engagement for biased and unbiased tweets and Table 7 reports OLS regressions on log-transformed engagement under two specifications: (1) log follower count as a covariate, and (2) user fixed effects, which absorb follower count alongside all other time-invariant account-level characteristics. After controlling for account reach, both bias dimensions predict significantly higher engagement across all three metrics ($p < .001$). For subjective bias, a biased tweet is associated with approximately 8–9% more likes, 4–5% more replies, and 18–36% more retweets relative to an unbiased tweet from the same account. For objective bias, effects are comparable: 3–9% more likes, 2–5% more replies, and ∼30% more retweets. The consistency across specifications suggests that follower count captures the primary confound. These results indicate that both affective and evidentiary forms of bias amplify engagement conditional on account reach: accounts that post objectively biased content tend to have smaller audiences, suppressing their raw engagement totals.

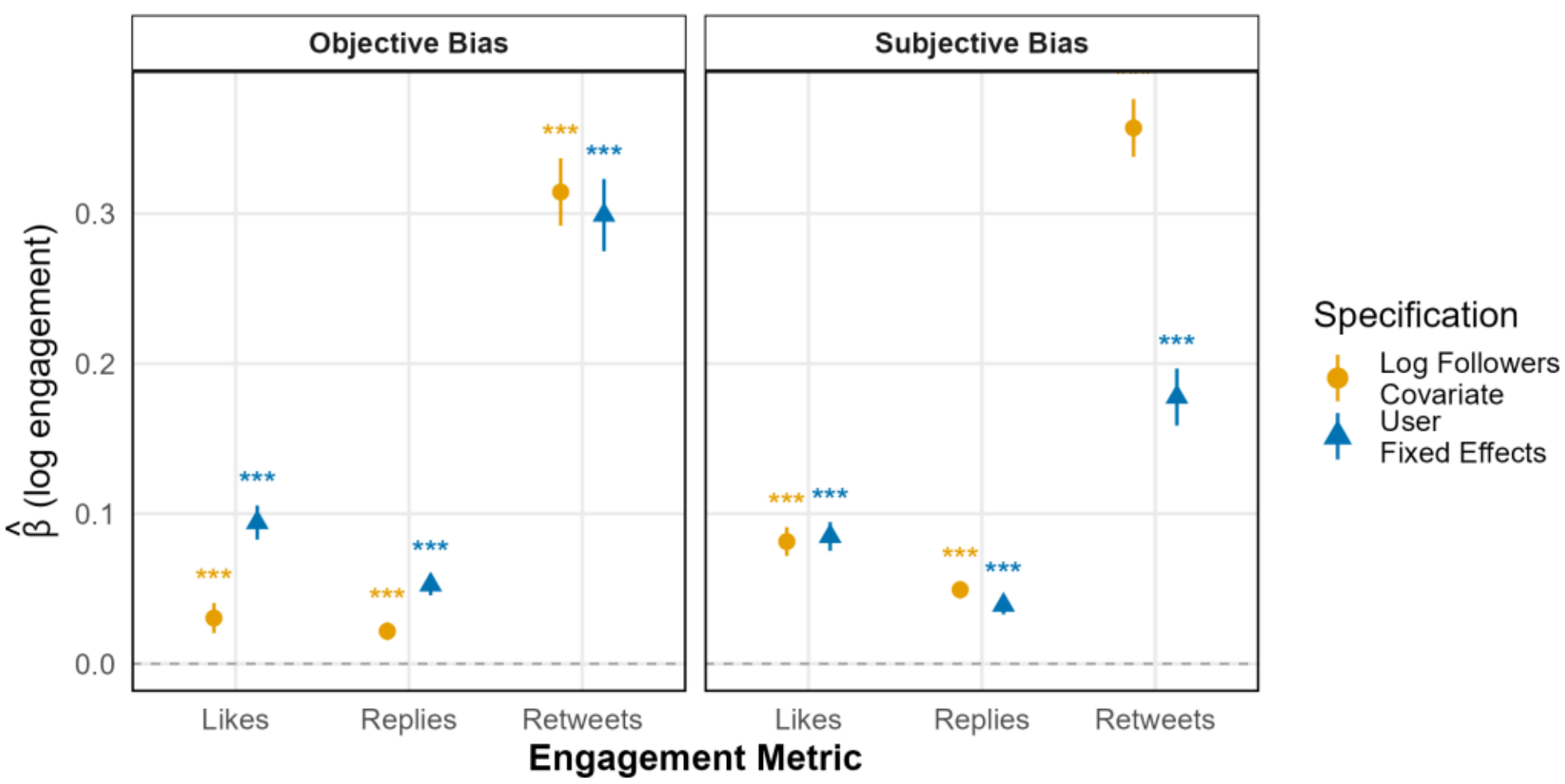

Figure 2: Regression coefficients ($\hat{\beta}$) for the effect of subjective and objective media bias on log-transformed engagement with the MediaSpin-in-the-Wild dataset, estimated under two specifications: log follower count as a covariate and user fixed effects. Points represent point estimates; whiskers are 95% confidence intervals with heteroskedasticity-robust standard errors. Both specifications consistently show positive, significant effects across all engagement metrics. $N = 180,758$ tweets from 819 users. $^{***}p < .001$.

Several mechanisms may explain why both forms of bias amplify engagement conditional on account reach. For subjective bias, the premium likely reflects the affective architecture of social platforms: emotionally charged language (e.g., "What You Don't Know Could Kill You"; Table 1) lowers the cognitive threshold for sharing by substituting visceral resonance for deliberation (Brady et al. 2017), while opinion framed as fact ("Euro Is Irreversible") invites agreement or disagreement that drives replies. For objective bias, the mechanism is less intuitive but consistent with work on information processing under uncertainty: content that strips source attribution or omits qualifying context removes the friction of evaluation can affect people's ability to endorse or amplify it (Sundar, Knobloch-Westerwick, and Hastall 2007). Slanted framing from a trusted source may similarly reduce the cognitive effort required to interpret an event, priming readers toward a ready-made reaction. Of course, affective and evidentiary distortions may exploit different cognitive shortcuts (Kahneman 2011), which is a hypothesis we will explore in future work.

## Discussion

The findings underscore the value of studying news framing as a *dynamic* process shaped by editorial revisions, rather than solely as a property of final published headlines. Editorial revisions provide empirical access to intermediate stages of meaning-making: the point at which agenda cues become interpretive frames, at which subjective and objective editorial strategies diverge, and at which textual micro-framings shape downstream diffusion. We highlight three broader implications.

**1. Edits clarify the transition from agenda-setting to framing.** Edits offer a methodological bridge: they show how agenda-setting (what is included) becomes framing (how it is interpreted) through concrete revision operations. The country-level analysis illustrates this directly—additions of national actors (e.g., the United States) and removals of regional labels (e.g., "Asia") transform an agenda cue into a frame that foregrounds a specific political actor. These revisions function as linguistic boundary-setting operations that delimit whose perspective is foregrounded and whose context is backgrounded, an aspect that prior computational work on labeling bias (Hamborg 2023) could not capture without process-level data.

**2. Subjective and objective bias reflect different editorial mechanisms.** Our analyses clarify that the two bias dimensions emerge through distinct editing operations. Subjective bias typically arises from additive revisions—introducing adjectives ("brutal"), intensifiers, or interpretive verbs ("warns," "slams")—that produce stronger emotional or adversarial frames. Objective bias is more frequently tied to subtrac-tive or structural changes: removing sources, deleting geographical context, or muting evidentiary qualifiers (Castillo-Campos, Becerra-Alonso, and Varona-Aramburu 2025). This finding also explains why automated classifiers detect subjective bias more reliably: objective bias requires identifying missing information rather than present linguistic markers.

**3. Editorial micro-framing interacts with platform dynamics.** Our engagement results show that after adjusting for account reach, both subjectively and objectively biased tweets attract significantly more likes and replies, yet, unadjusted the patterns are opposite to each other suggesting that these patterns are confounded by algorithmic recommendations and systemic differences in authors employing objective versus subjective triggers in framing news (Berger and Milkman 2012). In general, these associations suggest that editorially induced framing differences have measurable downstream consequences for news visibility.

The geographic asymmetries carry similar implications. Our finding that references to Western political actors are systematically foregrounded while regional labels are abstracted echoes long-standing concerns about uneven global news attention (Woo and Kim 2023; Amores, Arcila-Calderón, and González-de Garay 2020). When editorial revisions systematically de-specify references to conflict-affected populations, they risk contributing to "representational silencing"—the structural exclusion of particular actors from public discourse (Entman et al. 2004). Resources like MediaSpin can support media watchdog organizations in auditing whether editorial practices systematically disadvantage certain populations at moments when visibility is most consequential.

## Conclusion and Future Work

This study demonstrates how editorial revisions to news headlines can be systematically analyzed to detect linguis-tic and framing biases. MediaSpin shows that fine-grained edits—word additions, deletions, substitutions, and referential shifts—provide measurable indicators of subjective, objective, and political framing, advancing research on news framing and agenda-setting (Tewksbury and Scheufele 2019) by making the editorial process itself observable.

Future work could expand MediaSpin to additional languages and outlets, particularly non-Western media systems where representational dynamics carry the greatest stakes. A limitation of the current country-weighted scoring is that it cannot detect erasure for countries with very few mentions. Further work could integrate corpus-level attention measures with edit-level framing indicators to capture both presence and transformation. Multi-LLM annotation pipelines and larger-scale human validation would also strengthen reliability, particularly for objective categories requiring contextual inference.

## Ethical Considerations and Limitations

MediaSpin raises ethical and methodological considerations that warrant transparent discussion. We organize these around four themes: content sensitivity, inferential boundaries, scope constraints, and downstream risks.

**Content sensitivity.** The dataset contains headlines referencing gender, ethnicity, religion, violence, and conflict. These are retained to preserve ecological validity; they reflect what editors actually write and revise, but may be distressing depending on the use context. No personal information is included; all data derive from publicly available headlines. The engagement analysis (Section 6) draws on tweets from a non-representative sample of 819 consenting American X users, and behavioral patterns observed here should not be generalized to broader populations without replication across platforms, countries, and demographic groups.

**Inferential boundaries.** MediaSpin captures *what* changed in a headline, not *why* it changed. Edits may reflect routine copyediting, legal review, breaking-news updates, or evolving story clarity rather than ideological motivation.

**Scope constraints.** Several design choices limit the generalizability of the resource:

- *Outlet and language coverage.* The dataset includes five English-language outlets (Fox News, New York Times, Washington Post, Reuters, and Rebel News), largely based in the United States and Canada. Editorial norms, bias categories, and representational dynamics captured here reflect Anglophone Western media systems and should not be treated as globally representative. Categories such as sensationalism, mudslinging, or spin may manifest differently in non-English, non-Western, or state-controlled media contexts.
- *Temporal snapshot.* The dataset reflects the editorial practices documented within the time window of NewsEd-its (Spangher et al. 2022). Newsroom workflows evolve; future editing styles may differ in frequency, purpose, or linguistic form.
- *Annotation difficulty.* The thirteen bias categories vary in frequency and subtlety. Objective biases—omission, flawed logic, unsubstantiated claims—often require contextual knowledge beyond the headline itself, contributing to lower model accuracy and slightly reduced annotator agreement relative to subjective categories. Although our human-supervised LLM pipeline includes expert validation, GPT-based annotations may introduce inconsistencies, particularly for categories requiring inferential reasoning or domain knowledge.

**Downstream risks.** Models trained on MediaSpin could inadvertently amplify skewed patterns or be repurposed for persuasion, targeting, or political profiling. We explicitly discourage such uses. The intended applications are transparency research, media accountability, and bias auditing.

## Acknowledgments

This research was supported by the Singapore Ministry of Education through its MOE AcRF Tier 3 Grant (MOE-MOET32022-0001) and the Tier 1 programme (WBS A-8000231-01-00), and by the National Research Foundation, Singapore and Ministry of Communications and Information under its Online Trust and Safety (OTS) Research Programme Funding Initiative. Any opinions, findings and conclusions or recommendations expressed in this material are those of the authors and do not reflect the views of the funding organizations.

## Paper Checklist

1. For most authors...
   (a) Would answering this research question advance science without violating social contracts, such as violating privacy norms, perpetuating unfair profiling, exacerbating the socio-economic divide, or implying disrespect to societies or cultures?
   Yes. The dataset contains only publicly available news headlines, with no personal data or identifiable individuals. The research focuses on media framing and editorial changes, posing no privacy or profiling risks.
   (b) Do your main claims in the abstract and introduction accurately reflect the paper's contributions and scope?
   Yes. The abstract and introduction clearly state the dataset's content, annotation pipeline, analyses, and broader contributions without overclaiming.
   (c) Do you clarify how the proposed methodological approach is appropriate for the claims made?
   Yes. The paper explains why headline edits are a meaningful site for framing analysis and describes the annotation pipeline, validation process, and suitability of transformer-based models for bias classification.
   (d) Do you clarify what are possible artifacts in the data used, given population-specific distributions?
   Yes, in the Ethical Considerations section. The paper acknowledges outlet-level differences, annotation limitations (especially for objective bias), and uneven geographic representation, as well as categories with low sample frequency.
   (e) Did you describe the limitations of your work?
   Yes. Limitations are discussed in a dedicated section, including LLM annotation errors, restricted outlet coverage, temporal generalizability, and inability to infer editorial intent.
   (f) Did you discuss any potential negative societal impacts of your work?
   Yes. The Ethical Considerations section addresses risks such as misinterpretation of bias labels, amplification of sensitive content, and potential misuse of classifiers for political targeting.
   (g) Did you discuss any potential misuse of your work?
   Yes, in the Ethical Considerations section. The paper explicitly cautions against using the dataset for inferring journalist motives, persuasion-type applications, or profiling individuals or groups.
   (h) Did you describe steps taken to prevent or mitigate potential negative outcomes of the research, such as data and model documentation, data anonymization, responsible release, access control, and the reproducibility of findings?
   Yes. The dataset includes documentation, a datasheet, and an ethical-use statement; data release excludes personal information; models and annotations are described in detail for reproducibility.
   (i) Have you read the ethics review guidelines and ensured that your paper conforms to them?
   Yes. The paper follows the required guidelines on transparency, limitations, dataset documentation, and responsible use.
2. Additionally, if your study involves hypotheses testing...
   (a) Did you clearly state the assumptions underlying all theoretical results?
   Not applicable. The paper does not include hypothesis-testing or theoretical derivations.
   (b) Have you provided justifications for all theoretical results?
   Not applicable. No theoretical proofs or formal hypotheses are presented.
   (c) Did you discuss competing hypotheses or theories that might challenge or complement your theoretical results?
   Not applicable. The study is empirical and dataset-driven rather than theory-testing.
   (d) Have you considered alternative mechanisms or explanations that might account for the same outcomes observed in your study?
   Not applicable, as no causal or theoretical mechanisms are proposed.
   (e) Did you address potential biases or limitations in your theoretical framework?
   Not applicable. No theoretical framework requiring such analysis is presented.
   (f) Have you related your theoretical results to the existing literature in social science?
   Not applicable. The paper is empirical and methodological.
   (g) Did you discuss the implications of your theoretical results for policy, practice, or further research in the social science domain?
   Not applicable.
3. Additionally, if you are including theoretical proofs...
   (a) Did you state the full set of assumptions of all theoretical results?
   Not applicable. No theoretical proofs are included.
   (b) Did you include complete proofs of all theoretical results?
   Not applicable.
4. Additionally, if you ran machine learning experiments...
   (a) Did you include the code, data, and instructions needed to reproduce the main experimental results?
   Yes. The dataset is released via a Harvard Dataverse repository, and all model specifications, hyperparameters, and prompts are fully documented.
   (b) Did you specify all the training details (e.g., data splits, hyperparameters, how they were chosen)?
   Yes. Training hyperparameters, model choice (DeBERTa-v3-small), 70/30 splits, and the rationale for subjective vs. objective bias tasks are described.
   (c) Did you report error bars (e.g., with multiple seeds)?
   No. Models were run with a fixed seed and do not include error bars across multiple runs.

(d) Did you include the total amount of compute and the type of resources used?
Yes. The compute setup is described as part of the model section (single GPU environment).

(e) Do you justify how the proposed evaluation is sufficient and appropriate to the claims made?
Yes. Evaluation is aligned with the goal of demonstrating feasibility of automated bias classification rather than state-of-the-art optimization.

(f) Do you discuss what is "the cost" of misclassification and fault (in)tolerance?
Yes. The paper notes that objective bias categories are harder to detect and highlights the limitations of automated systems in editorial or policy contexts.

5. Additionally, if you are using existing assets or curating/releasing new assets...

(a) If your work uses existing assets, did you cite the creators?
Yes. NewsEdits and all prior datasets/tools are cited.

(b) Did you mention the license of the assets?
No explicit license is stated in the source dataset or in the text.

(c) Did you include any new assets in the supplemental material or as a URL?
Yes. MediaSpin is released via an open URL.

(d) Did you discuss whether and how consent was obtained from people whose data you're using/curating?
Yes. The dataset contains only public news headlines; no user-level or personal information is included, making additional consent unnecessary.

(e) Did you discuss whether the data contains personally identifiable information or offensive content?
Yes. The paper states that headlines may include sensitive or potentially offensive content as part of real-world reporting.

(f) If you are curating or releasing new datasets, did you discuss how you intend to make your datasets FAIR?
Yes. A FAIRification section is included describing findability, accessibility, interoperability, and reusability.

(g) If you are curating or releasing new datasets, did you create a Datasheet for the Dataset?
Yes. A full datasheet is included in the appendix.

6. Additionally, if you used crowdsourcing or conducted research with human subjects...

(a) Did you include the full text of instructions given to participants and screenshots?
Yes. The structured prompt for annotation (Figure 1) is fully included.

(b) Did you describe any potential participant risks, with mentions of Institutional Review Board (IRB) approvals?
Not applicable. No crowdsourcing or human-subject data beyond expert annotators was used.

(c) Did you include the estimated hourly wage paid to participants and the total amount spent on participant compensation?
Not applicable. No paid participants were involved.

(d) Did you discuss how data is stored, shared, and deidentified?
Yes. Only public, non-personal news text is stored; all identifying metadata is removed.

## Datasheet for the MediaSpin Dataset

This appendix provides a formal datasheet for the MediaSpin dataset, following recommended documentation standards for responsible dataset release. The dataset contains 78,910 headline pairs consisting of original and edited news headlines, annotated with thirteen media bias categories using a human-supervised LLM pipeline with expert validation. Please see a summary version in Table 6.

### Motivation

**Purpose.** MediaSpin was created to support the study of dynamic news framing by capturing post-publication edits in online news headlines. The dataset enables analyses of linguistic framing, editorial bias, and representational dynamics, as well as benchmarking of automated bias detection models.
**Creators.** The dataset was developed by the Preetika Verma and Kokil Jaidka as a supplement to the original dataset developed by Spangher et al. (2022).
**Funding.** No external funding sources are specified. The dataset was produced as part of the authors' academic research.

### Composition

**Instances.** Each instance is a headline pair (original + edited) with token-level added/removed words, POS tags, and 13 bias labels.
**Size.** 78,910 headline pairs selected from 376,944 edited pairs extracted from NewsEdits.
**Labels.** Thirteen media bias categories, including spin, unsubstantiated claims, opinion-as-fact, sensationalism, mudslinging, mind reading, slant, flawed logic, omission, omission of attribution, story choice, subjective adjectives, and word choice.
**Sources.** Five English-language outlets spanning ideological and credibility spectrums: Washington Post, Reuters, Fox News, New York Times, and Rebel News.
**Completeness.** Not exhaustive; curated from version histories of online articles.
**Potentially sensitive content.** Headlines may reference violence, political conflict, or sensitive issues.

### Collection Process

**Method.** Headline revisions were extracted from the NewsEdits dataset (Spangher et al. 2022). Token-level diffs were computed, cleaned, POS-tagged, and annotated using GPT-3.5-turbo with a structured prompt (Figure 1).
**Validation.** Two human annotators evaluated 509 samples;

agreement was 84.9%, with higher agreement for subjective bias categories.
**Timeframe.** Matches the revision window captured by NewsEdits (1.2M articles, 4.6M revisions).
**Ethics.** Only public news text is included; no personal data is collected.

### Preprocessing and Labeling

**Preprocessing.** Boilerplate phrases were removed, inserted/removed words extracted, POS tags assigned, and LLM-based labeling applied.
**Intermediate data.** Cleaned and annotated headline pairs constitute the dataset.
**Tools.** Standard Python NLP libraries and GPT-3.5-turbo; further details are documented in the paper.

### Uses

**Demonstrated uses.**

- Cross-national analysis of representational asymmetries (Figure 2)
- Bias classification using DeBERTa-v3 (Table 4)
- Behavioral analysis of biased headlines on X (Figure 3)

**Other uses.** Framing analysis, explainable NLP, bias auditing, media studies.
**Not recommended.** Inferring journalist intent; psychographic profiling; tasks outside the news domain without adaptation.

### Distribution

**Availability.** The Python and R package to annotate biases is open access, as is the dataset.

### Maintenance

**Maintainers.** Preetika Verma and Kokil Jaidka.
**Updates.** Possible future extensions (languages, outlets), but no guaranteed schedule.

## FAIRification of the MediaSpin Dataset

We follow the FAIR principles (Findable, Accessible, Interoperable, Reusable) to guide the responsible dissemination of the MediaSpin dataset.

### Findability

MediaSpin is assigned a stable URL at https://doi.org/10.7910/DVN/MOCQTZ. Metadata describing the dataset's structure, outlets, annotation schema, and collection methods accompanies the release. Each headline pair includes unique identifiers linking the original and edited versions, enabling granular reference and traceability.

### Accessibility

All files are provided in open, platform-independent formats (CSV and JSON), without requiring proprietary software. Access is unrestricted for research purposes, subject to ethical-use guidelines detailed in the accompanying documentation. No personal data is included; all headlines originate from publicly available news articles.

### Interoperability

The dataset uses standardized UTF-8 text encoding, consistent tokenization, and documented column names. Bias labels follow a fixed schema of thirteen media bias categories, enabling compatibility with existing tools for content analysis, NLP pipelines, and computational social science workflows. Country metadata is linked to standard geographic identifiers (e.g., GeoNames), facilitating integration with external geographic or sociopolitical datasets.

### Reusability

To maximize long-term usability, the dataset is distributed with a comprehensive description of the annotation pipeline, including the prompt used for LLM-based labeling (Figure 1), validation procedures, error patterns, and limitations. The license will allow research reuse (to be finalized post-review). Users are encouraged to cite the dataset to acknowledge the creators and to ensure the reproducibility of downstream modeling efforts. Detailed examples of annotated biases (Table 2), outlet metadata (Table 1), and model performance benchmarks (Table 4) support the immediate reuse of the dataset for benchmarking and methodological extensions.

## LLM Annotation prompts

Table 3 reports the annotation prompts used for the annotation of news headline pairs from the NewsEdits dataset.

## Regression results for Figure 2

Table 7 reports the OLS regression estimates of bias on log-transformed engagement with the MediaSpin-in-the-Wild dataset.

| Dataset Name | MediaSpin |
|---|---|
| **Creators** | Preetika Verma & Kokil Jaidka |
| **Motivation** | To analyze dynamic news framing by capturing post-publication edits to online news headlines, enabling fine-grained detection of framing bias and editorial intention across five major news outlets. |
| **Instances** | 78,910 headline pairs containing original and edited headlines, token-level edits (added/removed words), POS tags, and 13 media bias annotations. |
| **Source** | Curated from the NewsEdits dataset (Spangher et al. 2022), using version histories of online news articles. |
| **Outlets** | Washington Post, Reuters, Fox News, New York Times, Rebel (Table 1a). |
| **Labels** | 13 bias types: spin, unsubstantiated claims, opinion-as-fact, sensationalism, mudslinging, mind reading, slant, flawed logic, omission, omission of attribution, story choice, subjective adjectives, word choice. |
| **Annotation Method** | Human-supervised LLM annotation using GPT-3.5-turbo (Figure 1), plus expert validation over 509 samples with 84.9% agreement. |
| **Collection Mechanism** | Token-level diffing, text cleaning, POS tagging, LLM-based labeling, and human quality control. |
| **Temporal Coverage** | Aligned with the time period captured by the NewsEdits revision histories (1.2M articles, 4.6M revisions). |
| **Preprocessing** | Removal of boilerplate; extraction of inserted/removed tokens; POS tagging; LLM classification. |
| **Known Issues** | Low-frequency bias categories (e.g., flawed logic). Some LLM annotation errors, particularly in objective bias categories. |
| **Recommended Uses** | Bias classification, framing analysis, cross-national representational asymmetry, user-engagement analysis on X. |
| **Not Recommended For** | Inferring journalist intent, psychographic profiling, non-news-domain tasks without adaptation. |
| **Distribution** | The Python and R package to annotate biases is available at https://github.com/kokiljaidka/mediaspin. The dataset is available at https://doi.org/10.7910/DVN/MOCQTZ. |
| **Ethical Considerations** | Headlines may include sensitive political or violent content; annotators caution against misuse for political persuasion or targeted profiling. |

Table 6. Datasheet Summary for the MediaSpin Dataset.

Table 7. OLS regression estimates of bias on log-transformed engagement. All estimates are significant at $p < .001$ (***). 95% confidence intervals in brackets.

| Bias Type | Metric | Estimate | 95% CI | |
|---|---|---|---|---|
| *Panel A: Log Followers Covariate* | | | | |
| Subjective Bias | Likes | 0.081 | [0.072, 0.091] | *** |
| | Replies | 0.049 | [0.043, 0.055] | *** |
| | Retweets | 0.357 | [0.338, 0.376] | *** |
| Objective Bias | Likes | 0.030 | [0.020, 0.040] | *** |
| | Replies | 0.022 | [0.016, 0.028] | *** |
| | Retweets | 0.314 | [0.292, 0.337] | *** |
| *Panel B: User Fixed Effects* | | | | |
| Subjective Bias | Likes | 0.085 | [0.075, 0.094] | *** |
| | Replies | 0.039 | [0.033, 0.045] | *** |
| | Retweets | 0.178 | [0.159, 0.197] | *** |
| Objective Bias | Likes | 0.094 | [0.083, 0.105] | *** |
| | Replies | 0.053 | [0.046, 0.060] | *** |
| | Retweets | 0.299 | [0.275, 0.323] | *** |

You are a helpful assistant. You will be given a news headline and an edited version of the same headline. You will also be provided with a list of words that have been added to or removed from the original headline. Your goal is to label the words that have been added or removed based on their Part of Speech (POS). Additionally, you must analyze the changes to determine if they introduce or remove any of the following types of media bias. For each bias in the list, mention if it has been added, removed, or is not relevant to the headline.

**Types of Media Bias:**

- Spin (e.g., changing "protest" to "riot")
- Unsubstantiated Claims (e.g., adding "experts say" without providing evidence)
- Opinion Statements Presented as Fact (e.g., "The disastrous policy" instead of "The policy")
- Sensationalism/Emotionalism (e.g., "horrifying accident" instead of "accident")
- Mudslinging/Ad Hominem (e.g., "corrupt politician" instead of "politician")
- Mind Reading (e.g., 'He obviously didn't care' without evidence of feelings)
- Slant (e.g., highlighting only negative aspects of a story)
- Flawed Logic (e.g., "If A, then B" without proper justification)
- Bias by Omission (e.g., leaving out key details that support an alternative viewpoint)
- Omission of Source Attribution (e.g., making claims without citing sources)
- Bias by Story Choice and Placement (e.g., prioritizing negative news about a topic over positive news)
- Subjective Qualifying Adjectives (e.g., "the so-called expert" instead of "the expert")
- Word Choice (e.g., "freedom fighters" vs. "rebels")

**You will be provided input in the format:**

- Original Headline: [Provide the original headline here]
- Edited Headline: [Provide the edited headline here]
- Added words: Word_Added1, Word_Added2,..
- Removed words: Word_Removed1, Word_Removed2,..

**Your response must be of the format:**

- Words Added: Word_Added1 [POS], Word_Added2 [POS],..
- Words Removed: Word_Removed1 [POS], Word_Removed2 [POS],..

**Bias Analysis list:**

- 1. [type of bias] [Added/Removed/None]: The addition/removal of [specific word/phrase] introduces/removes this bias
- 2. [type of bias] [Added/Removed/None]: The addition/removal of [specific word/phrase] introduces/removes this bias
- 3. [type of bias] [Added/Removed/None]: The addition/removal of [specific word/phrase] introduces/removes this bias
- (...)
- N. [type of bias] [Added/Removed/None]: The addition/removal of [specific word/phrase] introduces/removes this bias.

Here is an example:

[INPUT EXAMPLE]
ALWAYS RESPOND IN THIS EXACT FORMAT.

Figure 3: Prompt for annotation by GPT-3.5.